\documentclass[sigconf]{acmart}

\usepackage{booktabs} 

\setcopyright{rightsretained}
\usepackage{caption}
\acmDOI{10.475/123_4}

\acmISBN{123-4567-24-567/08/06}

\acmConference[KDD'18]{KDD 2018}{August 2018}{London, UK}
\acmYear{2018}
\copyrightyear{2018}

\acmArticle{4}
\acmPrice{15.00}

\usepackage{subcaption}
\usepackage{graphicx}

\begin{document}
\title{Improving Long-Horizon Forecasts with Expectation-Biased LSTM Networks}

\author{Aya Abdelsalam Ismail}
\affiliation{%
  \institution{University of Maryland}
  \city{College Park}
  \state{USA}
}
\email{asalam@cs.umd.edu}

\author{Timothy Wood}
\affiliation{%
  \institution{University of Maryland}
  \city{College Park}
  \state{USA}
}
\email{twood1@terpmail.umd.edu}

\author{H\'{e}ctor Corrada Bravo}
\affiliation{%
  \institution{University of Maryland}
  \city{College Park}
  \state{USA}
}
\email{hcorrada@umiacs.umd.edu}

\renewcommand{\shortauthors}{A. Ismail, T. Wood and H. Bravo}

\begin{abstract}
State-of-the-art forecasting methods using Recurrent Neural Networks (RNN) based on Long-Short Term Memory (LSTM) cells have shown exceptional performance targeting short-horizon forecasts, e.g given a set of predictor features, forecast a target value for the next few time steps in the future. However, in many applications, the performance of these methods decays as the forecasting horizon extends beyond these few time steps. This paper aims to explore the challenges of long-horizon forecasting using LSTM networks. Here, we illustrate the long-horizon forecasting problem in datasets from neuroscience and energy supply management. We then propose expectation-biasing, an approach motivated by the literature of Dynamic Belief Networks, as a solution to improve long-horizon forecasting using LSTMs. We propose two LSTM architectures along with two methods for expectation biasing that significantly outperforms standard practice. 
\end{abstract}




\maketitle

\section{Introduction}

Modeling time varying data is a fundamental problem in Data Science with applications in a variety of fields such as medicine,  finance, economics, meteorology, and customer support center operations. Using models trained on time series data to forecast future values is a well studied area to which methods ranging from classical statical models such as 
Autoregressive integrated moving average (ARIMA), simple machine learning techniques such as support vector machines (SVM) have been applied. Of recent interest is the application of recurrent neural network (RNN), particularly networks based on state-of-the-art Long short-term memory (LSTMs)\cite{hochreiter1997long} cells. 

LSTM-based methods have shown great success in short-horizon forecasting, where forecasts are made for a small number of time steps beyond the last observations recorded in training data. Examples of this short-horizong forecasts are stock prediction \cite{kimoto1990stock},  recommender systems \cite{RRN} and ICU diagnosis \cite{lipton2015learning}. Unfortunately, these approaches can be lacking when long-horizon forecasts, where forecasts are made for a large number of time steps beyond the last recorded observations. The ability to capture temporal and causal aspects inherent in the data for longer time spans is missing in these models.

The following examples illustrate situations in which long-horizon forecasts are invaluable:

\par \textit{\textbf{Alzheimer's Prognosis}} A person with  Alzheimer's Disease (AD) that presents clear symptoms is usually accurately diagnosed by physicians. Unfortunately, by the time a person has developed clear symptoms of dementia or AD, treatment options are limited. There are no current treatments that provably cure or even slow progression of AD \cite{tagkey2017iii}. Clearly then, the value of identifying patients at early stages of the disease, or even before disease onset, is a key goal in utilizing effective preventative interventions.  This requires diagnosis years in advance; After all, it is much more challenging to capture patient changes in a long-horizon (e.g, predict that a cognitively normal patient will most likely develop dementia in 5 years) vs short-horizon (e.g, diagnose a patient who currently shows clear symptoms of dementia). As an illustration of the significance of this problem, the EuroPOND consortium created the TADPOLE challenge~\cite{tadpole} as a contest to predict AD prognosis over long-horizon forecasts.

\par \textit{\textbf{Energy Consumption}} Monitoring and controlling energy consumption is major issue in developed and developing countries. A fundamental requirement of power system operators is to maintain the balance between generation and load. An example of that is electricity consumption, for electrical power is a main energy form relied upon in all economic sectors all over the world. The prediction of electricity consumption, especially over long-horizons is a key component for successful long-term management (e.g. power flow) of the electrical grid. \cite{Arghira2013}

\par \textit{\textbf{Population growth}} 
Government policymakers and planners around the world use population projections to gauge future demand for food,
water, energy, and services, and to forecast future demographic characteristics. Population projections can alert policymakers to major trends that may affect economic development and help policymakers craft policies that can be adapted for various projection scenarios. The accuracy of population projections has been attracting more attention, driven by concerns about the possible long-term effects of aging, HIV/AIDS, and other demographic trends \cite{population}. To produce accurate projections,  the model must track changes in the attribute parameters such as the mortality rate, birth rate and health-care etc over time. This exaggerates the problem from just predicting population growth to predicting all the factors that affect the population growth either directly or indirectly.

\par \textit{\textbf{Long term economic consequence of political changes}} One of the largest political changes in the last decade was the Arab spring that swept the Middle East in early 2011 and dramatically altered the political landscape of the region. Autocratic regimes in Egypt, Libya, Tunisia, and Yemen were overthrown, giving hope to citizens towards a long-overdue process of democratic transition in the Arab world.  While the promise of democracy in the Arab transition countries was seen as the driving force in the uprisings, economic issues were an equally important factor. Political stability is very difficult, if not impossible, to achieve if the economy is in disarray \cite{khan2014economic}, therefore studying the economical effects of the Arab spring on the Middle East became a very interesting problem for economists. Long term and short term effects are very different in this case and to be able to capture each appropriately, changes in parameter trends over time must be taken into consideration.

\begin{figure}[H]
  \centering
  \includegraphics[width=2.7in]{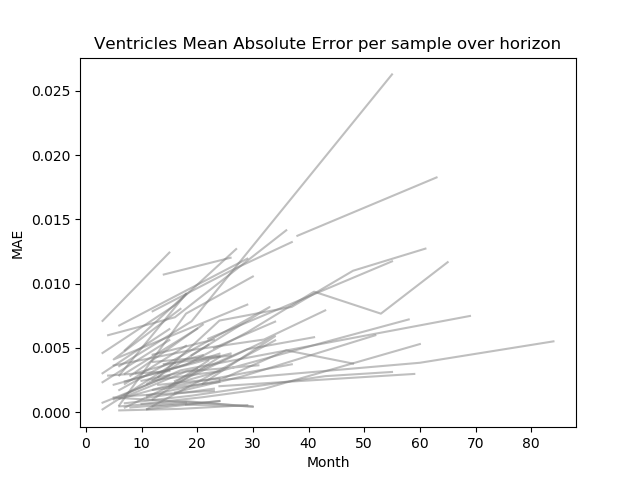}
  \caption{Change in MAE for ventricle volume forecasting over time. Each line represents the mean absolute forecasting error of a single participant}
  \label{fig:fig1}

\end{figure}

\begin{figure}[H]
  \centering
  \includegraphics[width=2.7in]{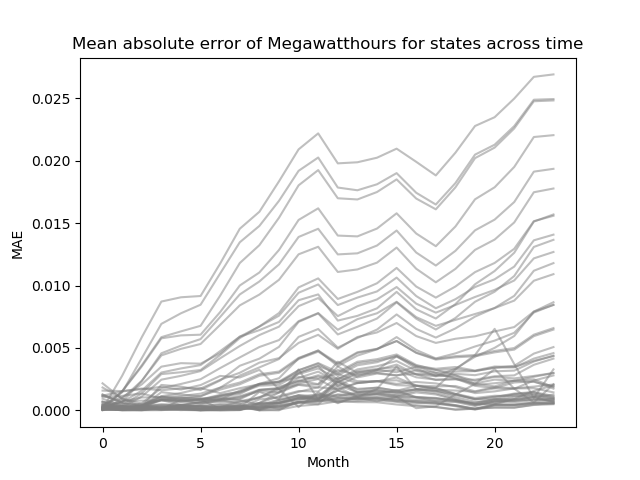}
  \caption{Change in MAE for electricity consumption over time. Each line represents the mean absolute forecasting error of a single state}
  \label{fig:fig2}
\end{figure}

In this paper we focus on the first two examples: we look at forecasting brain ventricular volume as a biomarker for neurodegenerative disease progression \cite{nestor2008ventricular}; we also look at forecasting electrical energy consumption in the United States.c To illustrate the problem of long-horizon forecasting, we applied a state of the art Long Short-Term Memory (LSTM) \cite{hochreiter1997long} model datasets in these application areas (dataset details below). This is depicted in Figures \ref{fig:fig1} and \ref{fig:fig2}, where each shows the change in mean absolute forecast error over the forecasting horizon when predicting ventricular volume and electricity consumption respectively. In both cases, the LSTM model achieves good results for short-horizon forecasts. However, as the forecasting horizon increases, accuracy worsens significantly.

\noindent 
\newline
\newline
\newline

To improve prediction accuracy, we can instead build conditional multivariate models that incorporate relevant features for prediction. For instance, in the Alzheimer's case we could include other physiological measurements for each subject, or in the electricity case, we could incorporate metereological and economic indicators that could affect electricity consumption. A challenge faced when using these multivariate models is that we need predictive features at all time points in order to make forecasts, but data from future time points are obviously not available. This scheme may yet improve prediction accuracy if we assume that values for these predictive features persist over the forecasting horizon. This assumption may hold for short-horizon forecasts, but is obviously violated in long forecasting horizons. We could at this point train a multivariate LSTM network to forecast predictive features along with the target forecast of interest but the difficulty of training LSTM networks increases substantially as the number of predictive features increases, making this approach prohibitive in many cases.

This paper proposes, implements, and evaluates a scheme to inject bias into LSTM networks in a manner that significantly improves the accuracy of their forecasts over long-term horizons. Our contributions are as follows:
\begin{itemize}
\item  \textbf{Introducing long-horizon forecasting problem in LSTMs}. We introduce and discuss the problem of LSTM architectures in creating forecasts over long-horizons and illustrate the problem in two distinct important application areas. 

\item \textbf{Propose methods to incorporate bias into LSTM models that significantly reduce the the long-horizon forecasting problem} To the best of our knowledge this is the first paper to address the long-horizon forecasting problem in LSTMs. That is, this is the first multivariate model that captures changes over a long period of time on a large scale. 

\item \textbf{Empirical evaluation of long-horizon forecasts} We show cases in which long-horizon problem exists, and show that introducing model bias to LSTM improves the accuracy and outperforms state-of-the-art models.
\end{itemize}

\section{Background}

\subsection{LSTM}
\begin{figure}[H]
\centering
\includegraphics[width=3in]{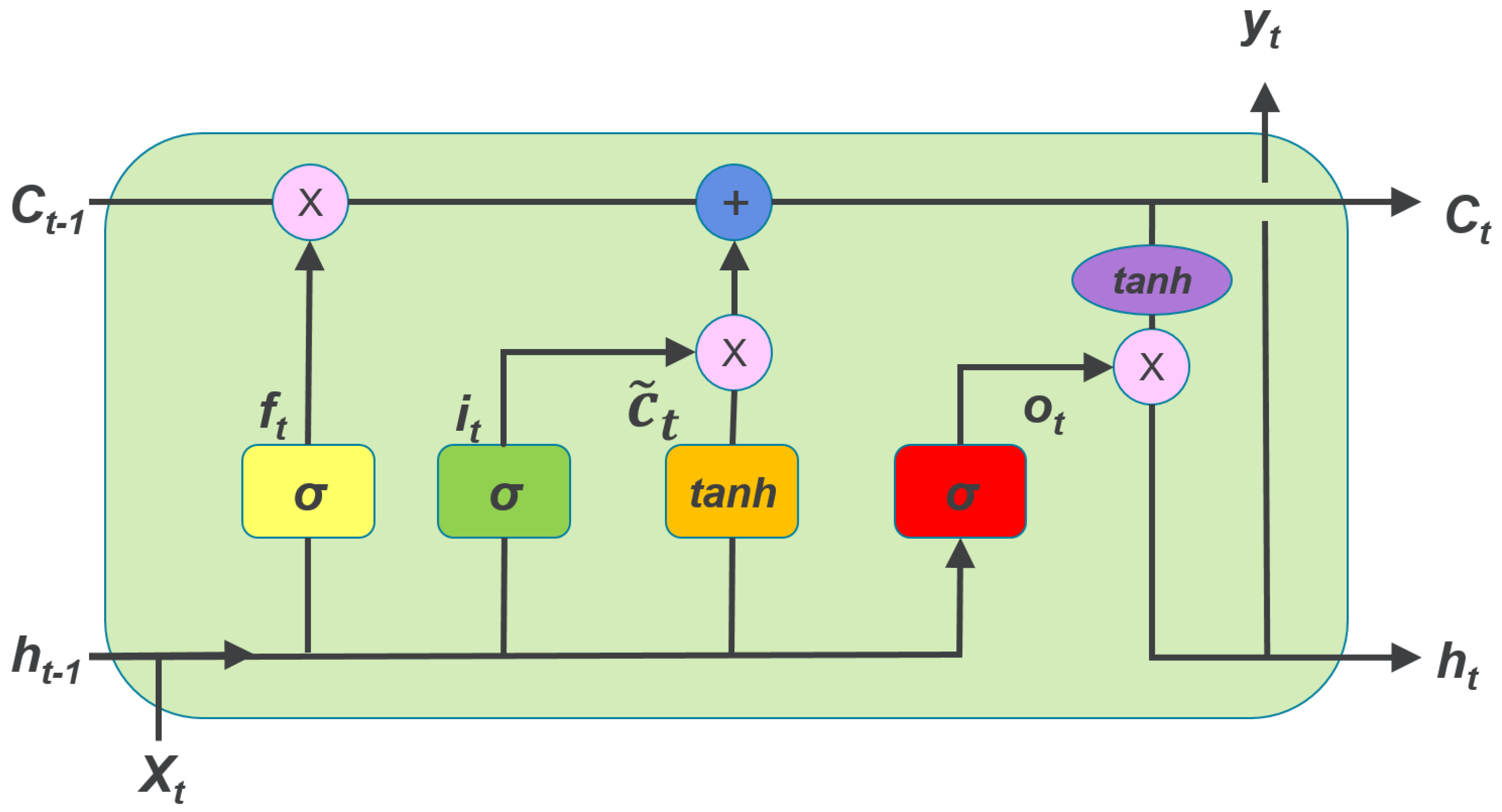}
  \caption{Long Short-term Memory Neural Network}
  \label{fig:lstm}
\end{figure}

A popular choice for forecasting are recurrent neural networks (RNNs) based on Long Short-Term Memory (LSTM) \cite{hochreiter1997long} cells. Since LSTMs incorporate memory units that explicitly allow the network to learn when to "forget" previous hidden states
and when to update hidden states given new information, they have been utilized effectively for sequences or temporally based data. The LSTM cell is shown in figure \ref{fig:lstm}.

The state updates satisfy the following operations:
 
\begin{equation}
\begin{split}
		& \mathbf i_t = \sigma \left(\mathbf W_{xi}\mathbf x_t + \mathbf W_{hi}\mathbf h_{t-1} + \mathbf b_i\right)\\
		& \mathbf f_t= \sigma\left(\mathbf W_{xf}\mathbf x_t + \mathbf W_{hf}\mathbf h_{t-1} + \mathbf b_f\right)\\
		& \mathbf o_t= \sigma \left(\mathbf W_{xo}\mathbf x_t+ \mathbf W_{ho}\mathbf h_{t-1} + \mathbf b_o\right)\\
       & \mathbf {\tilde{c_t}} = \tanh \left(\mathbf  W_{xc}\mathbf x_t + \mathbf W_{hc}\mathbf h_{t-1} + \mathbf b_c \right)\\
		& \mathbf c_t = \mathbf f_t \odot \mathbf  c_{t-1}+  \mathbf i_t \odot \mathbf {\tilde{c_t}}  \\
		& \mathbf h_t = \tanh \left(\mathbf o_t \odot \mathbf c_t\right) 
\end{split}
\label{eq:1}
\end{equation}

Here $\sigma$ is the logistic sigmoid function and $\odot $ is the Hadamard (element-wise) product;
$\mathbf W_i, \mathbf W_f, \mathbf W_o, \mathbf W_c$ are recurrent weight matrices;  $\mathbf b_i, \mathbf b_f, \mathbf b_o, \mathbf b_c$ are the bias terms. In addition to a hidden unit $\mathbf h_t$ , the LSTM includes
an input gate $\mathbf i_t$ , forget gate $\mathbf f_t$ , output gate $\mathbf o_t$ , input modulation gate $\mathbf {\tilde{c_t}}$ , and memory cell $\mathbf c_t$. 

While other variations of LSTM are commonly employed, including Gated recurrent unit GRU \cite{gru}, No Forget Gate (NFG), No Peepholes (NP), No Input Activation Function (NIAF), and others, a comparison between different variations of LSTM\cite{greff2017lstm} showed that while computationally cheaper, these variations did not improve upon the standard LSTM architecture significantly. Since we are proposing a general purpose long-horizon forecasting model we will use the standard LSTM shown in figure \ref{fig:lstm} as the basis of the RNNs we implement in this paper.

\subsection{Dynamic Network Models} \label{DNM}
Dynamic Network Models (DNM) \cite{dagum1992dynamic} are a probabilistic method to define multivariate conditional models for time series. Introduced in 1992 they extend static belief network to support time series data by including temporal dependencies between representations of a static belief network across times. 

\begin{figure}[H]
  \centering
  \includegraphics[width=2in]{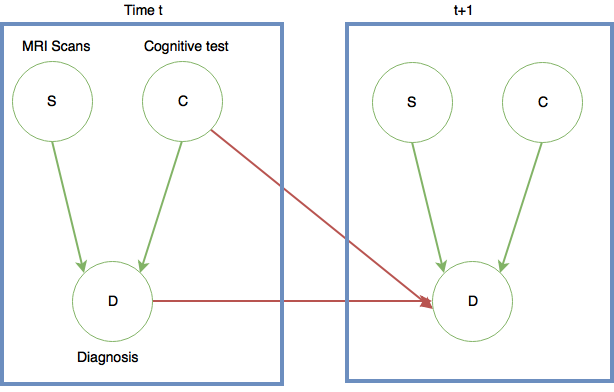}
  \caption{Simple Dynamic Network Model}
  \label{fig:DNM}
\end{figure}

Consider a simple application in Alzheimer's diagnosis where you have two predictor features, say a summary of an MRI scan and the result of a cognitive test, and a target of interest to be forecasted, for example, Alzheimer's diagnosis. Figure \ref{fig:DNM} shows a DNM where the nodes represent the state of these variables at time $t$, and edges indicate conditional dependence structure within predictors and the target variable. There are two types of dependencies, contemporaneous dependencies between variables at the same time point. In our example, arcs between nodes with types MRI scans, cognitive tests and diagnosis at time $t$  indicate the dependence of diagnosis on predictor features at specific time $t$. Non-contemporaneous dependencies are indicated by edges between nodes at different time points, for example, time $t$ and $t+1$.

These models are usually defined by parametric models of the conditional probability distributions specified by the contemporaneous and non-contemporaneus structure described by the network as illustrated above. As such, these methods present two issues: network structure has to be either pre-specified or learned from data (belief structure learning is a challenging problem), and estimation of the parameters of the conditional probability models themselves which is usually done via maximum-likelihood \cite{Figueredo:2009dg} via updates to parameters as observations are processed over time arrive. Once parameters are estimated, the conditional-probability distributions of the model may be re-evaluated. In the example mentioned above the diagnosis at time $t$ depends on the MRI scans and cognitive tests at time $t$, denoted by $Q[d_t|s_t,c_t]$. In addition, the diagnosis at time $t$ depends on the diagnosis and cognitive tests at time $t-1$, denoted by $R[d_t|d_{t-1},c_{t-1}]$. Hence the prediction of diagnosis is given by :

\begin{equation}
\begin{aligned}
Pr\left[d_t|s_t,c_t,d_{t-1},c_{t-1};\alpha \right]=
&\alpha Q \left[d_t|s_t,c_t\right] +\\ 
& \left(1-\alpha \right) R \left[d_t|d_{t-1},c_{t-1}\right]
\label{eq:13}
\end{aligned}
\end{equation}

Where $\alpha$ is the likelihood that the diagnosis predicted from the information time $t-1$ is correct, and $ \left(1-\alpha \right)$ is the likehood that diagnosis predicted from the information time $t$ is correct. Here, $\alpha$ is calculated using maximum-likelihood after each new observation and hence the next prediction uses the new $\alpha$ such that the model considers the changes in contemporaneous dependencies and  non-contemporaneous dependencies over time. We will use a similar concept in our work but use a non-parametric LSTM networks to make forecasts, which circumvents the structure learning issue discussed above, but presents shortcomings over the DBN idea for long-horizon forecasts.

In order to perform forecasts using DBNs we would solve 

$$
\arg \max_{d_t} E_{s_t,c_t,d_{t-1},c_{t-1}} Pr \left[ d_t|s_t,c_t,d_{t-1},c_{t-1};\alpha \right],
$$

where $E_{X}(f)$ corresponds to the \textit{expectation} of function $f$ over $X$. In this case, $f$ is the probability model defined by the DBN and $X$ the predictor features. Methods like the EM algorithm, based on the parameteric assumption of the model, are used to solve this optimization problem. This leads to a natural interpretation where forecasts are, in a sense, biased using the expectation of predictor features. Our proposed methods in this paper are motivated by this observation. We extend non-parametric LSTM models by adding additional input features that capture the expectation of predictor features so forecasts are therefored biased in a sense similar to that of DBNs. 


\section{Models}
In this section, we describe two different forecasting schemes: a multivariate LSTM with a single output shown in Figure \ref{fig:Model1}, and a multivariate LSTM with multiple outputs shown in Figure \ref{fig:Model2}. We also propose two methods to compute expectation bias and show how to incorporate it expectation bias into each of these two architectures.

\subsection{Multivariate LSTM with single output and bias}\label{Multivariate_single_output}
\begin{figure}[H]
\centering
\includegraphics[width=2in]{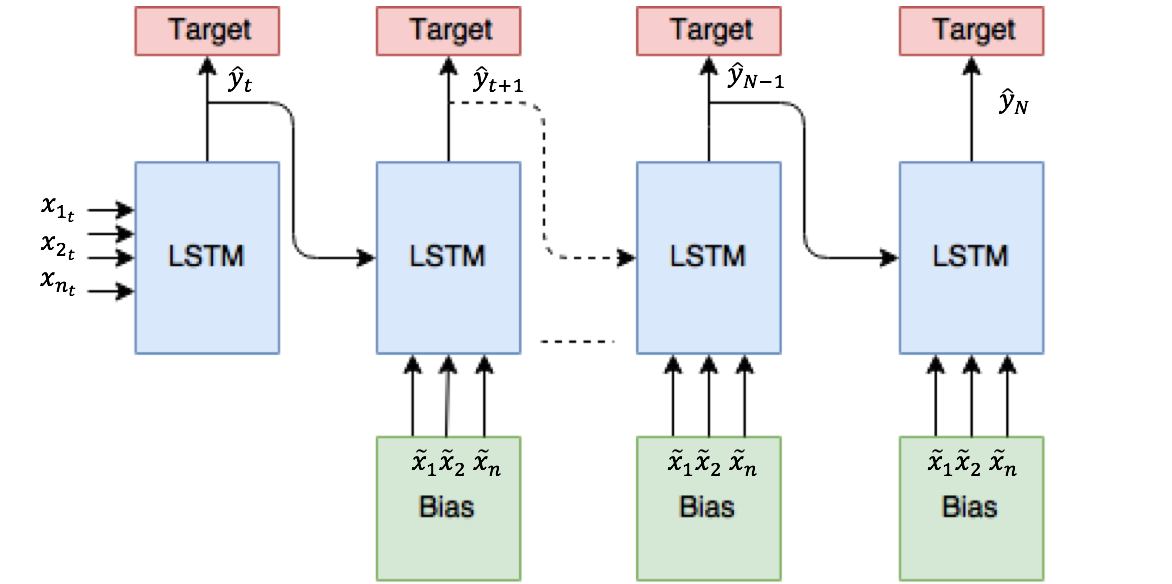}
  \caption{ Multivariate LSTM with 4 features and a single output. The output of LSTM at time $t$ is the input of the next LSTM at time $t+1$. The 3 remaining features are calculated using the bias function. N represents some time in the future. \footnotesize }
  \label{fig:Model1}
\end{figure}

The first approach, model 1 (Figure \ref{fig:Model1}), describes Multivariate LSTM Recurrent Neural Networks. The input to the first LSTM are the observed predictor features, but for all future LSTMs the input is the expectation bias term $e_{i,t}$ for time $t$ and the value of output from the previous LSTM. We introduce a new term  $\tilde x_{i,t} $:

\begin{equation}
\begin{aligned}
\tilde x_{i,t}  = \left\{
                \begin{array}{ll}
                  x_{i,t} \text{ at } t=0\\
                  e_{i,t}  \text{ at } t \neq 0
                \end{array}
              \right.
  \label{eq:8}
\end{aligned}
\end{equation}
$e_{i,t}$ is the output of the expectation bias function at time $t$ for feature $i$. We will define how to compute this in Section \ref{section_bias}. The resulting model is:
\begin{equation}
\begin{aligned}
\hat{ y}_{t+1} = \text{LSTM} (\tilde x_{1,t}, \tilde x_{2,t},\ldots , \tilde x_{n,t} , \  \hat{y_t})\label{eq:9}
\end{aligned}
\end{equation}
where $n$ is the number of features and $\hat{y}$ is the predicted value.
The optimization objective is to minimize the difference between the predicted output and the actual output, i.e.
\begin{equation}
\begin{aligned}
\text{minimize }  \left( \text{ loss} \left(\hat{y},y\right)\right)
\label{eq:10}
\end{aligned}
\end{equation}

\subsection{Multivariate LSTM with multiple outputs and bias}\label{Multivariate_multiple_output}
\begin{figure}[H]
\centering
\includegraphics[width=2in]{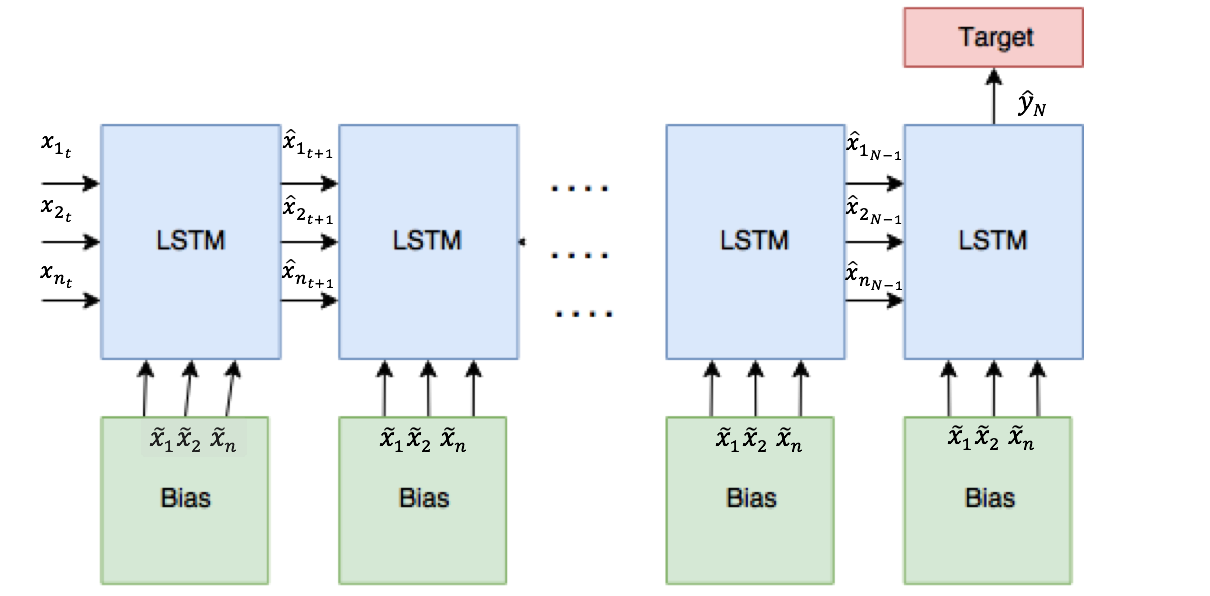}
   \caption{Multivariate LSTM where the outputs of the previous LSTM are concatenated with bias as inputs to the next.\footnotesize }
  \label{fig:Model2}
\end{figure}

Figure \ref{fig:Model2} shows a multivariate LSTM recurrent neural network, where the outputs of an LSTM are concatenated with inputs from a bias function then passed as inputs to the next LSTM. As such, all features are predicted via the model for each time step.

Each feature is represented as a concatenation of predictor features and the expectation bias:

\begin{equation}
\begin{aligned}
\tilde{x}_{i,t}  \coloneqq  \left[ x_{i,t},e_{i,t} \right]\label{eq:11}
\end{aligned}
\end{equation}

where, as before, $e_{i,t}$ is the output of the bias function for time $t$ and $x_{i,t}$ is feature $i$ at time step $t$. Now, we have the following model:
\begin{equation}
\begin{aligned}
\hat{ x}_{1,t+1},\hat{ x}_{2,t+1},\ldots ,\hat{ x}_{n,t
+1} = \text{ LSTM} (\tilde x_{1,t}, \tilde x_{2,t},\ldots , \tilde x_{n,t})\label{eq:12}
\end{aligned}
\end{equation}
We define a loss function based on the predicted value and the observed values across all features, i.e.

\begin{equation}
\begin{aligned}
\text{loss}(\hat{x},x) = \sum_{i=1}^{n} \alpha_i \cdot \text{ loss} \left( \hat{x}_i,x_i \right) 
\label{eq:13}
\end{aligned}
\end{equation}

Here, $\hat{x}_i$ is the predicted value of feature $x_i$ with $\alpha_i \in \left[0,1\right]$ a hyper-parameter which indicates the importance of each feature such that $\sum_{i=1}^{n}\alpha_i = 1 $. 
\newline
\newline



The target is finally predicted as a function of the outputs LSTM, i.e
\begin{equation}
\begin{aligned}
\hat{y}_{t+1} = \text{ f} (\hat{x}_{1,{t+1}},\ldots, \hat{x}_{n,{t+1}})\label{eq:12}
\end{aligned}
\end{equation}
Similar to the multivariate LSTM in model 1, the latest observations are taken as input and predictions are generated. The objective function is given by

\begin{equation}
\begin{aligned}
\text{minimize } \left( \alpha_ \cdot \text{ loss} \left( \hat{x},x \right) +
\left(1-\alpha\right) \cdot \text{ loss} \left(\hat{y},y\right)\right)
\label{eq:13}
\end{aligned}
\end{equation}

where $\alpha$ is another hyper-parameter that weighs prediction error between target and the loss of the LSTM in predicting feature values.

\subsection{Bias}\label{section_bias}
We motivated the idea of incorporating expectation bias to improve long-horizon forecasts from dynamic network models described in section \ref{DNM}. In this section we show how to compute the expectation bias terms included in the two models defined above using empirical expectation estimates of predictor features $x_i$. Here, input to the bias function is the features $x_{1,t},\ldots,x_{n,t}$ and time $t$. We propose two methods for bias calculation below.

\subsubsection{\textbf{Population Averaging:}}
In this method, expectation bias is calculated based on stationary population averages of predictor features $x_i$ using the following equation:

\begin{equation}
\begin{aligned}
e_{i,t} = \beta \left(t\right) x_{i} + \left(1- \beta \left(t\right) \right) \mu_i
\label{eq:14}
\end{aligned}
\end{equation}

where, $\mu_i$ is the popoulation average of predictor feature $x_i$ across observations $N$:

\begin{equation}
\begin{aligned}
\mu_i = \frac{1}{N} \sum_{j=0}^{j=N} x_{ij}
\label{eq:15}
\end{aligned}
\end{equation}

$\beta \left(t\right)$ is a time varying function defined by the user depending on the dataset. This function can differ in complexity from simply being $\beta \left(t\right) =  \frac{1}{t}$  to a more complex equation. However, $\beta \left(t\right) \in \left[0,1\right]$ where $\beta \left(t\right) = 1$ at $t=1$ and  $\beta \left(t\right) = 0$ at $t \rightarrow \infty$. 

\subsubsection{\textbf{ Clustering:}} We can extend the above model to better capture heterogeneity across the $N$ observations in a dataset. Here, we group observations together, specifically using the K-Means clustering algorithm \cite{hartigan1979algorithm} and use empirical expectation estimates within clusters to define empirical bias. 

Determining the number of clusters in a data set is not straight forward, since the correct choice of $K$ is often ambiguous. There are several methods to for choosing $K$ including, the elbow method, X-means clustering, silhouette method and cross-validation. In our implementation, the silhouette method \cite{rousseeuw1987silhouettes} was used to get an optimal value for $K$, and different $Ks$ around the given optimal were also considered, i.e if initial $K=3$ then  $K=2$ and  $K=4$ were tested to evaluate the sensitivity of our method to the number of clusters.

Euclidean distance was used to assign samples to cluster center. Let $c_j$ denote the center of cluster $j$, $x_i$ is the value of feature $i$ for the sample, and $n$ is the number of features per sample; the distance is then calculated by the following equation:
\begin{equation}
\begin{aligned}
\sum_{i=1}^{i=n} \sum_{j=1}^{j=K} || x_{i} -c_{j_{i}}||^2
\label{eq:16}
\end{aligned}
\end{equation}
Thus, a given sample is assigned to its closest center. The cluster centers are calculated during training to reduce computational overhead at prediction. The assigned cluster center is the bias value

\begin{equation}
\begin{aligned}
e_{i,t} = c_{j_{i}}
\label{eq:14}
\end{aligned}
\end{equation}



\section{Experiments} \label{exp}
We show cases where long-horizon problem exists and how unbiased LSTMs are not able to provide accurate predictions for such problems. We compare the performance of biased and unbiased in models 1 and 2, where unbiased baselines are defined as follows.

\begin{figure}[H]
\centering
\includegraphics[width=2in]{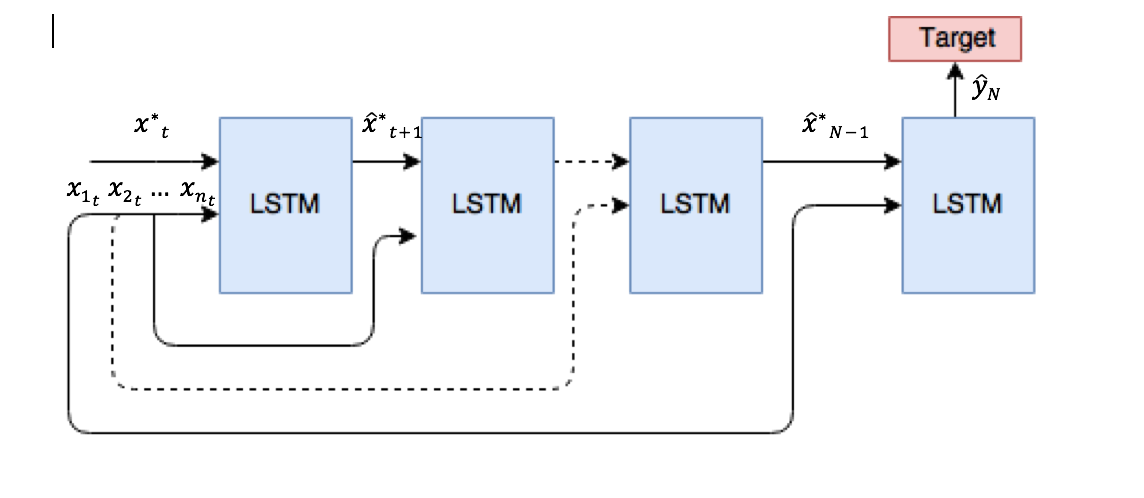}
  \caption{Unbiased multivariate LSTM with single output}
  \label{fig:Model1_BL}
\end{figure}

\textbf{Baseline 1: \textit{Unbiased multivariate LSTM with a multiple output.}}
Model 1 described in section \ref{Multivariate_single_output} is compared with unbiased multivariate LSTM shown in figure \ref{fig:Model1_BL}. The output of LSTM  at time $t$ is $\hat{x}^*_{t+1}$, while its inputs are  $\hat{x}^*_{t}$ and the initial value of all other features $x_1 , x_2 , \ldots , x_n$ at time $t=1$. 

\begin{figure}[H]
\centering
\includegraphics[width=2in]{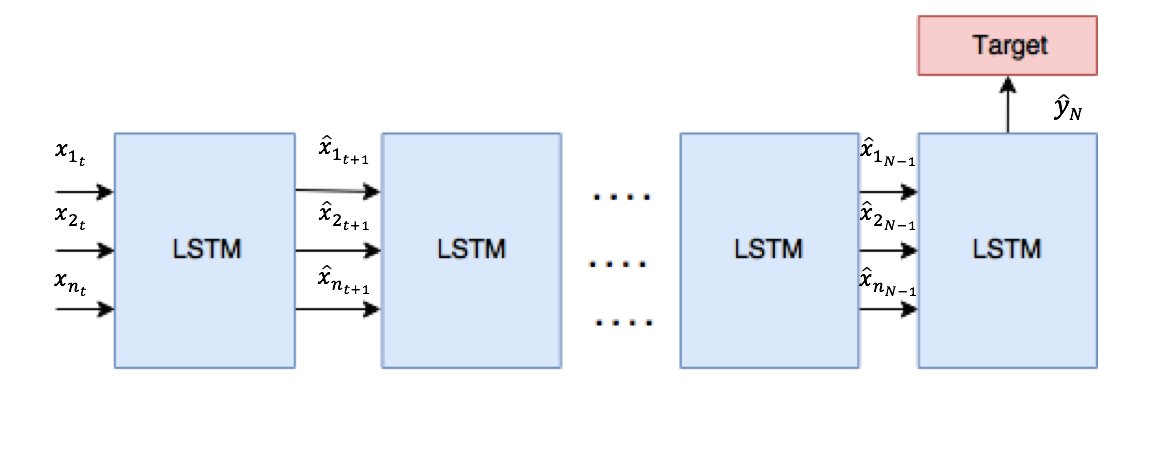}
  \caption{Unbiased multivariate LSTM with multiple output}
  \label{fig:Model2_BL}
\end{figure}

\noindent\textbf{Baseline 2: \textit{ Unbiased multivariate LSTM with a multiple output}}
Model 2 described in section \ref{Multivariate_multiple_output} is compared with unbiased multivariate LSTM shown in figure \ref{fig:Model2_BL}. The outputs of LSTM  at time $t$ are the inputs to the next LSTM at time $t+1$. 

\noindent\textbf{Accuracy:} The mean absolute error (MAE) at every timestep is used to measure model accuracy. MAE at timestep $t$ is given by
\begin{equation}
\begin{aligned}
MAE_{t} = \frac{1}{M} \sum_{i=0}^{i=M}  |\hat{y}_{i_{t}} - y_{i_{t}}|
\label{eq:mAE}
\end{aligned}
\end{equation}
where $M$ is the number of observations acquired by the time the forecasts are evaluated, $y_{i_{t}}$ is the actual value at that time $t$ and $\hat{y}_{i_{t}}$ is the predicted value at time t.

\noindent\textbf{Evaluation:} In order to study the effectiveness of expectation bias on long-horizon forecasting, we used two publicly available datasets to evaluate the models. The first dataset is from the Alzheimer's Disease Neuroimaging Initiative (ADNI) \footnote{Data used in preparation of this article were obtained from the Alzheimer's Disease Neuroimaging Initiative (ADNI) database (adni.loni.usc.edu). As such, the investigators within the ADNI contributed to the design and implementation of ADNI and/or provided data but did not participate in analysis or writing of this report. A complete listing of ADNI investigators can be found at:  \href{http://adni.loni.usc.edu/wp-content/uploads/how_to_apply/ADNI_Acknowledgement_List.pdf}{ADNI Acknowledgement List}}, this dataset was used to predict the change in ventricular volume over time. We are particularly interested in forecasting the ventricular volume since there is a link between ventricular enlargement and Alzheimer's disease (AD) progression \cite{nestor2008ventricular} \cite{ott2009relationship}. The second dataset is monthly electricity consumption across the United States for individual states. The dataset is obtained from Energy Information Administration  (EIA) \cite{outlook2008energy} and NOAA's National Centers for Environmental Information (NCEI) \cite{NOAA}. Here, the goal is to forecast amount of electricity in Megawatt hours consumed by each state in the United States per month.

Details of the datasets, setup and results will be given in the sections bellow. 

\subsection{Ventricular forecasting}\label{VF}
\subsubsection*{\textbf{Data Description}}

\begin{figure}[H]
\centering
\includegraphics[width=2in]{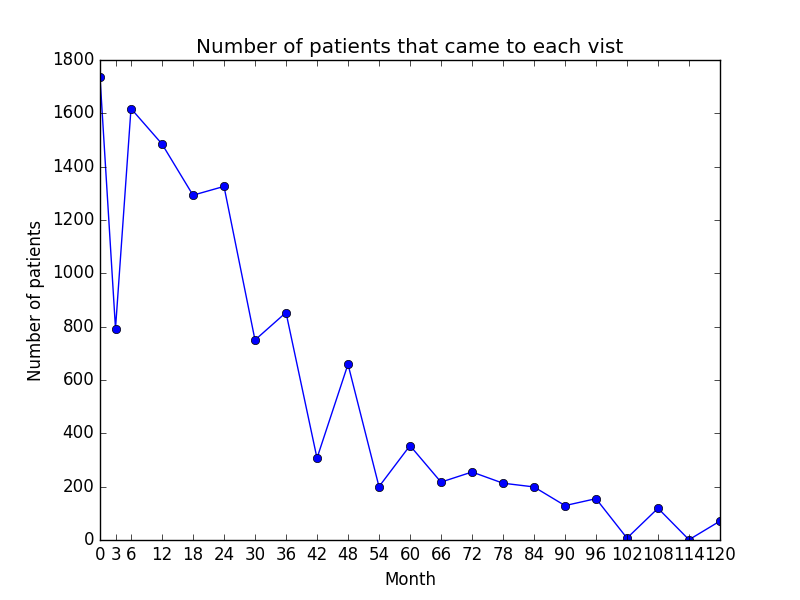}
  \caption{Number of subject that came to their bi-yearly visit}
  \label{fig:peoplepervisit}
\end{figure}

We are using the ADNI dataset published as part of the Tadpole challenge \cite{tadpole}, which consists of 1628 subjects that are either cognitively normal, stable mild cognitive impairment, early mild cognitive impairment, late mild cognitive impairment or have Alzheimer disease. This data was collected over a 10 year period, and measurements were typically taken every six months. Since subjects are allowed to enter and leave the trial at any point the number of visits per subject was different. Figure \ref{fig:peoplepervisit} shows the number subjects observed at each of the follow-up visits. 

The data consists of 1835 features that represent the following biomarkers:
\begin{itemize}
\item Neuropsychological tests administered by a clinical expert.
\item MRI metrics measuring brain structural integrity.
\item FDG Positron Emission Tomography (PET) measuring cell metabolism, where cells affected by AD show reduced metabolism.
\item AV45 and AV1451 PET measuring a specific protein (amyloid-beta and tau respectively) load in the brain.
\item Diffusion Tensor Imaging (DTI) inferring microstructural parameters related to cells and axons.
\item Cerebrospinal fluid (CSF) Biomarkers.
\item Others including diagnosis, genetic and demographic information.
\end{itemize}
\subsubsection*{\textbf{Data preparation}}
In order to prepare data for forecasting the following steps were followed:
\begin{itemize}
\item \textbf{Make all observations have same amount of time between them} Since all visits are 6 months apart except for the second visit, which is 3 months after the initial visit, all data from visit 2 was discarded.
\item \textbf{Transform time series into supervised learning problem} This was done by making feature data at time t-1 the feature targets, and Ventricle value at time t the prediction target.
\item \textbf{Feature Scaling} Feature standardization was used, which makes the values of each feature in the data have zero-mean (when subtracting the mean in the numerator) and unit-variance (dividing by the standard deviation). The general method of calculation is to determine the distribution mean and standard deviation for each feature, and standardize from there. 

\item \textbf{Missing data} We analyzed missing data and divided each observation into two categories: missing at random or not missing at random. For example if one exam is only done once at the first visit for all subjects then it is not missing at random, and in this case we insert the correct value of that exam for future time points. For data missing at random, Hot Deck was used for imputation.

\item \textbf{Feature selection} We used a random forest \cite{breiman2001random} to select features with highest cross entropy reduction. The package used was the standard randomForest package in R.
\end{itemize}
\subsubsection*{\textbf{ Setup}}
The setup is similar to that of the TADPOLE Challenge \cite{tadpole}. Data was divided into two datasets: a training dataset that contains a set of measurements for every individual that has provided data to ADNI in at least two separate visits, and a test data set. Individuals in the test dataset are not used when forecasting or training/building training models. The goal is to make month-by-month forecasts for normalized ventricular volume of each individual in the test dataset for a period of 5 years. Training Dataset was then divided into two sub-dataset: training and validation to facilitate hyperparameter estimation.

Our LSTM was implemented using TensorFlow \cite{DBLP:journals/corr/AbadiABBCCCDDDG16}.
We train each LSTM for 500 epochs using  Adam: A Method for Stochastic Optimization \cite{DBLP:journals/corr/KingmaB14}. We used variable length LSTM since the number of visits per subject is not constant. Our final networks use 2 hidden layers and 64 memory cells per layer with a learning rate 0.0003. These architectures are also chosen based on validation performance. For population averaging bias $\beta (t)$ is defined as:

\begin{equation}
\begin{aligned}
\beta (t)  = \left\{
                \begin{array}{ll}
                  1 \text{ at } t<20\\
                  0  \text{ Otherwise}
                \end{array}
              \right.
  \label{eq:8}
\end{aligned}
\end{equation}

Where 20 was chosen as it is the maximum number of possible visits in the training dataset. For clustering bias, the performance of Models when $K=2$, $K=3$ and $K=4$ were compared.

\begin{figure}[H]
\centering
\includegraphics[width=2in]{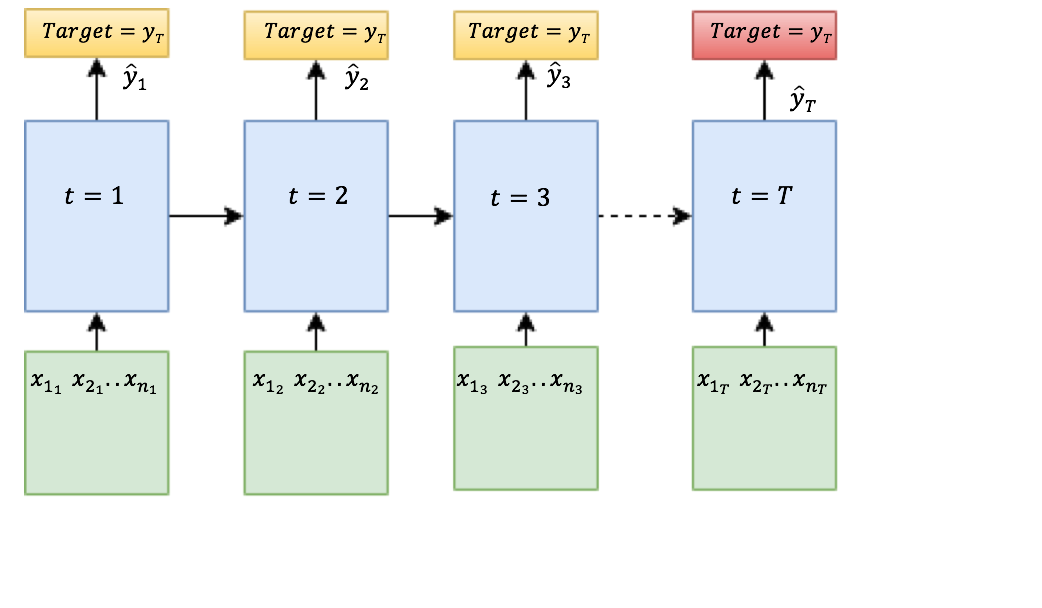}
  \caption{Single unrolled LSTM }
  \label{fig:UNROLLED}
\end{figure}

Sequential target replication  \cite{lee2015deeply}  was used for training, i.e if the LSTM has $T$ timesteps then the error at that timestep $t=2$ is $loss\left( \hat{y}_2 ,y_T \right)$. Our loss is a combination of the final loss and the average of the losses over all steps:

\begin{equation}
\begin{aligned}
\alpha .\frac{1}{T-1} \sum_{t=1}^{t=T-1} \text {loss} \left( \hat{y}_t ,y_T \right) +  (1-\alpha) . \text {loss} \left( \hat{y}_T ,y_T \right)
  \label{eq:8}
\end{aligned}
\end{equation}
$\alpha$ is a hyper-parameter that indicates the importance of each time step. Figure \ref{fig:UNROLLED} shows an unrolled LSTM with target replication. Each box represents a timestep, and the target for all timesteps is equal to that of the last timestep $y_T$. This technique  teaches the network to pass information across many sequence steps in order to affect the output. Target replication showed promising results when applied to a medical dataset in
Lipton, Zachary C., et al. \cite{lipton2015learning}. 

\subsubsection*{\textbf{Results}}

\begin{figure}[H]
\centering
\includegraphics[width=2in]{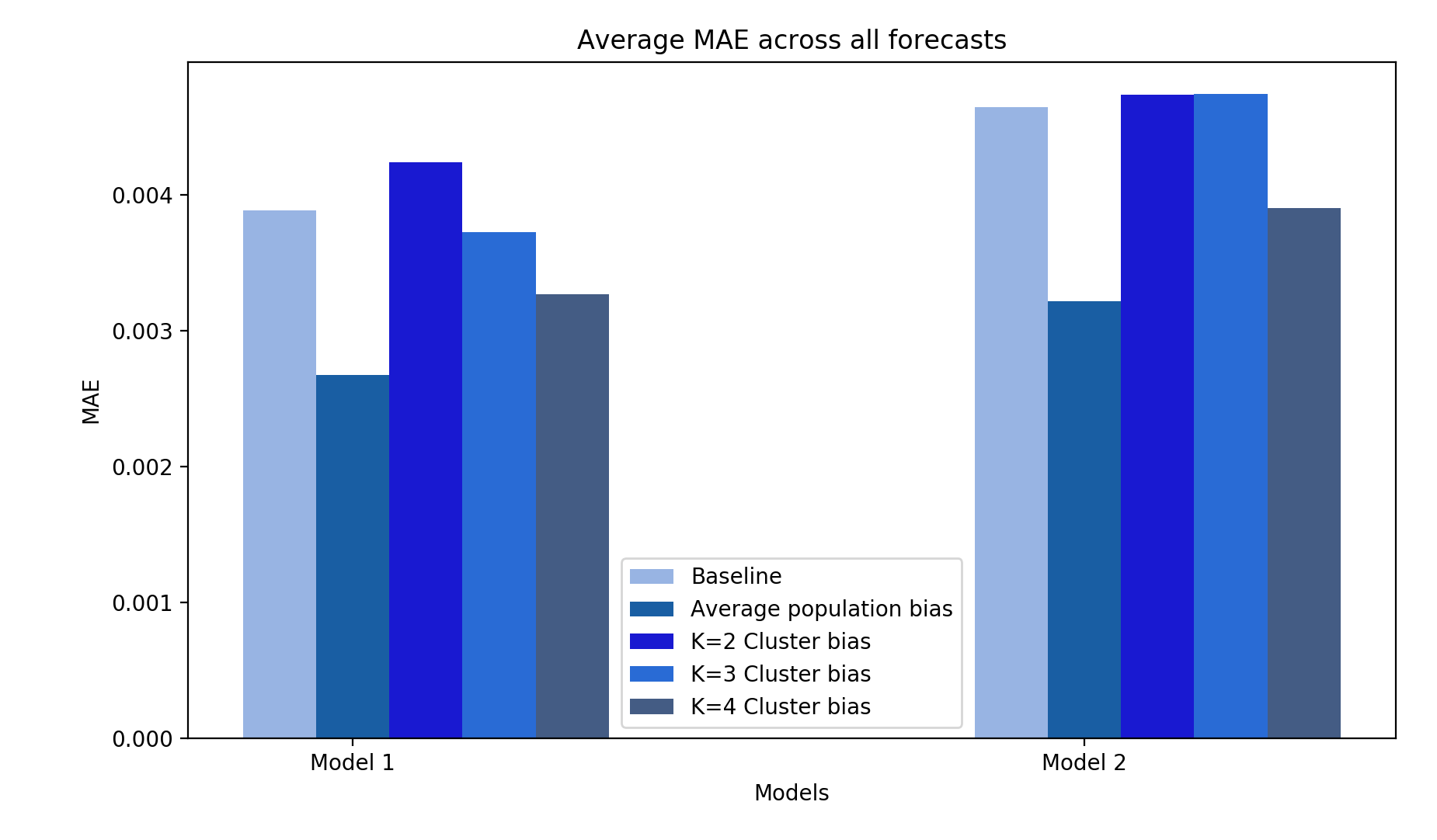}
  \caption{Comparing average MAE of biased and unbiased models}
  \label{fig:DS1_MAE}
\end{figure}

Figure \ref{fig:DS1_MAE} summaries the performance of unbiased models in comparison with biased models. The bars represent the average of the mean absolute error across all forecasts. The plot shows that in most cases biased model do better or as well as unbiased models. Figure also shows that the value of $K$ in biased cluster model can affect the performance significantly.

\subsection{Electricity forecasting}
\subsubsection*{\textbf{Data Description}}
The dataset contains 4 features; electricity consumption in Megawatt Hours, population count, price of Kilowatt Hours, and average temperature. Price of electricity was obtained from the Energy Information Administration (EIA), \cite{outlook2008energy} while average temperature of that state was obtained from NOAA's National Centers for Environmental Information (NCEI) \cite{NOAA}. Data from January 2007 till October 2017 monthly for every state and the District of Columbia was available in the dataset.
\subsubsection*{\textbf{Data preparation}}
\begin{figure} [H]
\centering
\begin{subfigure}[b] {0.55\textwidth}
\centering
   \includegraphics[width=0.4\linewidth]{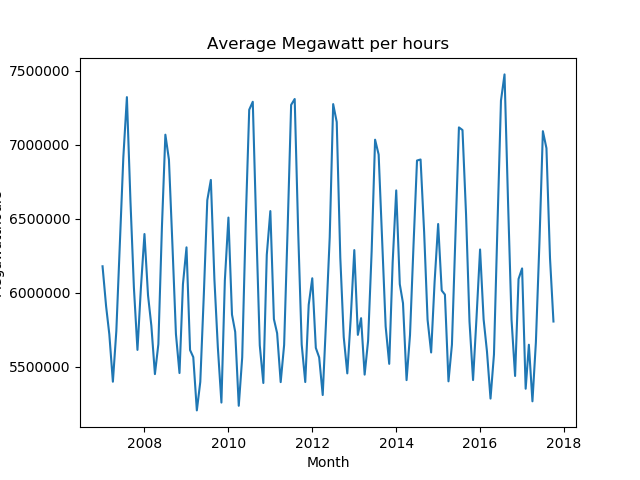}
\end{subfigure}

\begin{subfigure}[b]{0.55\textwidth}
\centering
   \includegraphics[width=0.4\linewidth]{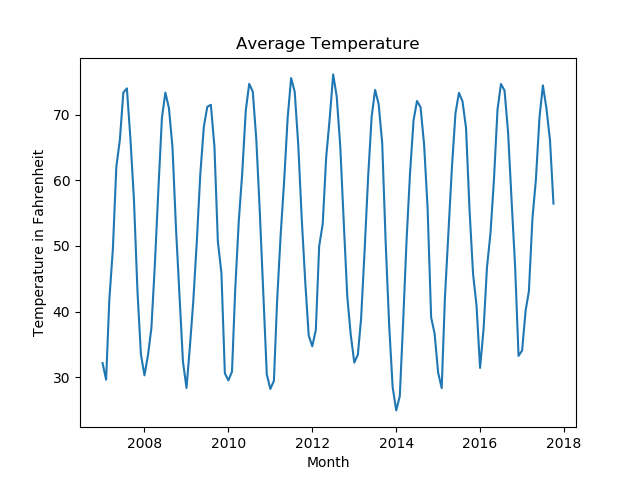}
\end{subfigure}

\begin{subfigure}[b]{0.55\textwidth}
\centering
   \includegraphics[width=0.4\linewidth]{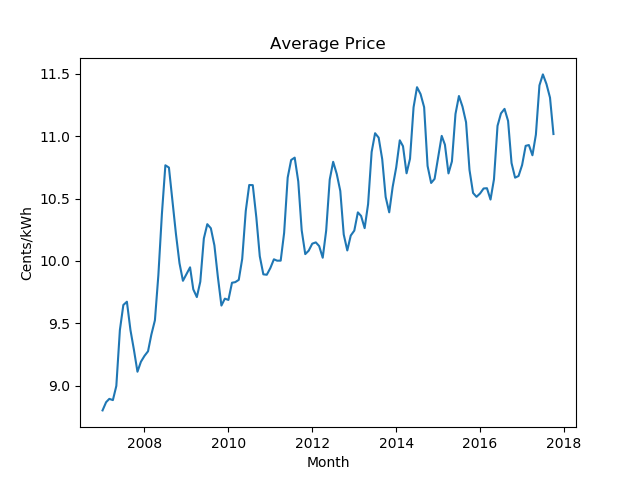}
\end{subfigure}

\caption[]{Seasonality is shown when plotting the average of different features in the dataset}
  \label{fig:DS2_TS} 
\end{figure}

One problem with this dataset is seasonality - the presence of variations that occur at specific regular intervals less than a year. Figure \ref{fig:DS2_TS} is a plot of average electricity consumption in Megawatt Hours, temperature and price across all states; all three features show seasonality. We transformed the time series by removing dependencies between observations, so the difference between observations was used as features instead of the the observation value. Features where then scaled using the same method described in section \ref{VF}.

\subsubsection*{\textbf{Setup}}
Models were trained on data from January 2007 through October 2015. We then made month-by-month forecasts of normalized Megawatt Hours of electricity consumed by each state from November 2015 through October 2017. A 2 layer LSTM with 64 memory cells per layer and learning rate of 0.0001 was trained with 500 epochs using Adam. For population averaging bias $\beta (t)  = \frac{1}{t}$ was used, and for clustering bias the performance of Models when $K=2$, $K=3$ and $K=4$ were compared. Sequential target replication was not used for this dataset.


\subsubsection*{\textbf{Results}}
\textbf{Model 1:}
Baseline for model 1 for this dataset is shown in Figure \ref{fig:fig2}, unbiased LSTM was able to accurately predict the first month but as forecasting horizon increased the accuracy of the baseline model decreased significantly for all states (observations).

\begin{figure}[H]
\centering
\includegraphics[width=2in]{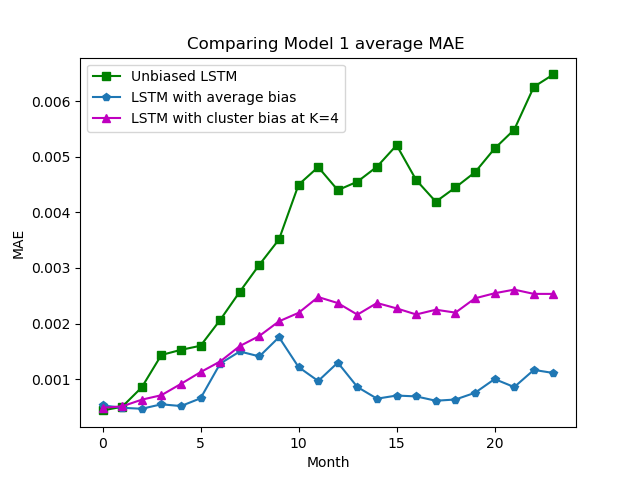}
  \caption{Average mean absolute error across all states between Model 1 baseline with population average and cluster biased LSTM at $K=4$}
  \label{fig:M1D2_Graph1}
\end{figure}

Figure \ref{fig:M1D2_Graph1} compares unbiased LSTM (baseline) with biased LSTM using population average and clustering  at $K=4$, the plot shows that adding expectation bias significantly improved LSTM performance. Although the long-horizon forecasting problem did not disappear completely, i.e., accuracy still decreases in the expectation biased models as the forecasting horizing increases, however the effect is significantly reduced.

\begin{figure}[H]
\centering
\includegraphics[width=2in]{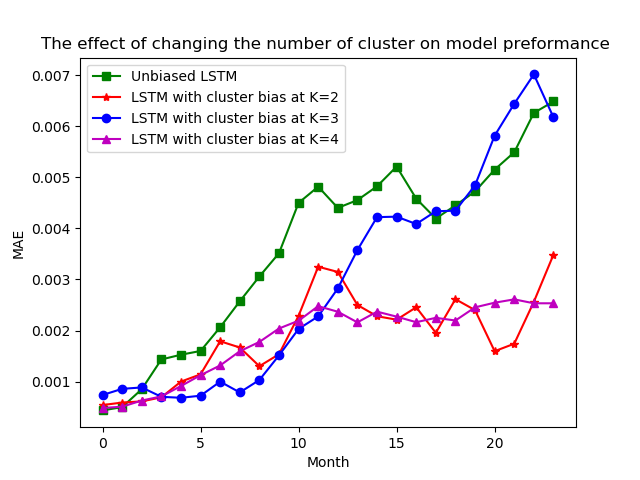}
  \caption{Comparing the effect of changing the number of cluster centers in a biased LSTM}
  \label{fig:M1D2_Graph2}
\end{figure}

Figure \ref{fig:M1D2_Graph2} shows the effect of changing the number of cluster centers on biased LSTM. For models with clusters $K=2$ and $K=4$ expectation biased significantly improves LSTM performance across the entire horizon, the model with $K=3$ outperforms the baseline at the beginning but then its performance deteriorates quickly. This shows that like many machine learning techniques the choice of the number of cluster will affect the performance of the model.

\textbf{Model 2:}

\begin{figure}[H]
\centering
9\includegraphics[width=2in]{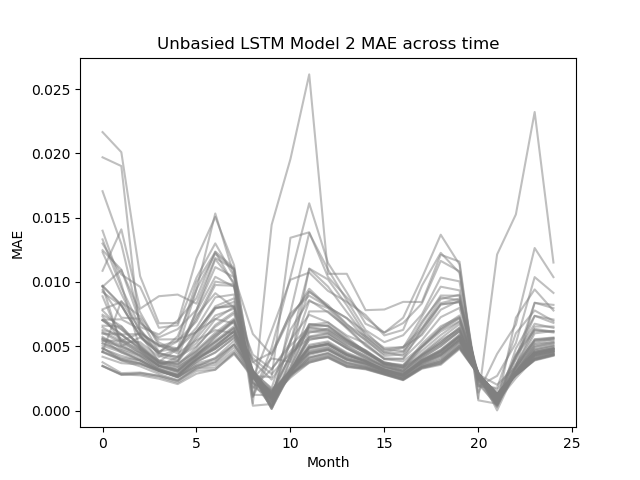}
  \caption{Mean absolute for each state in an unbiased multivariate LSTM with multiple outputs}
  \label{fig:DS1M2_BL}
\end{figure}

Baseline (unbiased) performance for model 2 is shown in Figure \ref{fig:DS1M2_BL}. In this case, the long-horizon forecasting problem is not as severe since the number of features predicted by the multiple output LSTM is small (4) and the number of time points was relatively large and thus the LSTM was able to learn to predict the features fairly well. This shows that in cases where the number of features is limited, one might consider forecasting all of the features in order to prevent the long-horizon forecasting problem.

Since this dataset for model 2 did not show horizon problem adding bias did not help increase the accuracy of LSTM. Note however, that performance of expectation biased model 1 is better than unbiased model 2.


  


\section{Conclusion and Future work}
We described long-horizon forecasting problem in LSTM and introduced two different LSTM forecasting architectures that incorporate expectation bias, an idea motivated by the Dynamic Belief Network literature. We have shown that severe long-horizon forecasting accuracy decay can be significantly improved by expanding LSTMs to incorporate expectation bias as proposed here.

Perhaps a key point to our results is that our models are relatively simple, and are therefore amenable to training with moderate number of observations. Unlike methods that would address long-horizon forecasting by possibly increasing the depth of LSTM networks, our method of bias can improve error without necessarily singificantly increasing the cost of training models over those used in standard practice. Furthermore, our expectation bias models scale well (especially the linear scaling of Model 1) as the dimensionality of the problem applications increases.

In future work, we plan to conduct further in-depth studies on architecture changes to the state model of LSTM cells to, analogous to the long and short term memory implemented by these cells, capture both short and long-term horizon forecasts. We also plan to explore other expectation biasing methods. For instance, using the EM algorithm instead of K-means to obtain expectation estimates, and using Autoregressive integrated moving average (ARIMA) based methods. In addition to these ideas, we will gauge the effectiveness of expectation bias on deep learning models that struggle to maintain accuracy in the distant future.

\bibliographystyle{ACM-Reference-Format}
\bibliography{example_paper}


\begin{thebibliography}{23}


\ifx \showCODEN    \undefined \def \showCODEN     #1{\unskip}     \fi
\ifx \showDOI      \undefined \def \showDOI       #1{#1}\fi
\ifx \showISBNx    \undefined \def \showISBNx     #1{\unskip}     \fi
\ifx \showISBNxiii \undefined \def \showISBNxiii  #1{\unskip}     \fi
\ifx \showISSN     \undefined \def \showISSN      #1{\unskip}     \fi
\ifx \showLCCN     \undefined \def \showLCCN      #1{\unskip}     \fi
\ifx \shownote     \undefined \def \shownote      #1{#1}          \fi
\ifx \showarticletitle \undefined \def \showarticletitle #1{#1}   \fi
\ifx \showURL      \undefined \def \showURL       {\relax}        \fi
\providecommand\bibfield[2]{#2}
\providecommand\bibinfo[2]{#2}
\providecommand\natexlab[1]{#1}
\providecommand\showeprint[2][]{arXiv:#2}

\bibitem[\protect\citeauthoryear{Arghira, Ploix,
  F{\u{a}}g{\u{a}}r{\u{a}}{\c{s}}an, and Iliescu}{Arghira
  et~al\mbox{.}}{2013}]%
        {Arghira2013}
\bibfield{author}{\bibinfo{person}{Nicoleta Arghira},
  \bibinfo{person}{St{\'e}phane Ploix}, \bibinfo{person}{Ioana
  F{\u{a}}g{\u{a}}r{\u{a}}{\c{s}}an}, {and} \bibinfo{person}{Sergiu~Stelian
  Iliescu}.} \bibinfo{year}{2013}\natexlab{}.
\newblock \bibinfo{booktitle}{\emph{Forecasting Energy Consumption in
  Dwellings}}.
\newblock \bibinfo{publisher}{Springer Berlin Heidelberg},
  \bibinfo{address}{Berlin, Heidelberg}, \bibinfo{pages}{251--264}.
\newblock
\showISBNx{978-3-642-32548-9}
\urldef\tempurl%
\url{https://doi.org/10.1007/978-3-642-32548-9_18}
\showDOI{\tempurl}


\bibitem[\protect\citeauthoryear{Breiman}{Breiman}{2001}]%
        {breiman2001random}
\bibfield{author}{\bibinfo{person}{Leo Breiman}.}
  \bibinfo{year}{2001}\natexlab{}.
\newblock \showarticletitle{Random forests}.
\newblock \bibinfo{journal}{\emph{Machine learning}} \bibinfo{volume}{45},
  \bibinfo{number}{1} (\bibinfo{year}{2001}), \bibinfo{pages}{5--32}.
\newblock


\bibitem[\protect\citeauthoryear{Brian~O'Neill}{Brian~O'Neill}{[n. d.]}]%
        {population}
\bibfield{author}{\bibinfo{person}{Deborah~Balk Brian~O'Neill}.}
  \bibinfo{year}{[n. d.]}\natexlab{}.
\newblock \showarticletitle{Understanding and using population projections}. In
  \bibinfo{booktitle}{\emph{Population reference bureau}}.
\newblock


\bibitem[\protect\citeauthoryear{Chung, Gulcehre, Cho, and Bengio}{Chung
  et~al\mbox{.}}{2014}]%
        {gru}
\bibfield{author}{\bibinfo{person}{Junyoung Chung}, \bibinfo{person}{Caglar
  Gulcehre}, \bibinfo{person}{KyungHyun Cho}, {and} \bibinfo{person}{Yoshua
  Bengio}.} \bibinfo{year}{2014}\natexlab{}.
\newblock \showarticletitle{Empirical evaluation of gated recurrent neural
  networks on sequence modeling}.
\newblock \bibinfo{journal}{\emph{arXiv preprint arXiv:1412.3555}}
  (\bibinfo{year}{2014}).
\newblock


\bibitem[\protect\citeauthoryear{Dagum, Galper, and Horvitz}{Dagum
  et~al\mbox{.}}{1992}]%
        {dagum1992dynamic}
\bibfield{author}{\bibinfo{person}{Paul Dagum}, \bibinfo{person}{Adam Galper},
  {and} \bibinfo{person}{Eric Horvitz}.} \bibinfo{year}{1992}\natexlab{}.
\newblock \showarticletitle{Dynamic network models for forecasting}.
\newblock In \bibinfo{booktitle}{\emph{Uncertainty in Artificial Intelligence,
  1992}}. \bibinfo{publisher}{Elsevier}, \bibinfo{pages}{41--48}.
\newblock


\bibitem[\protect\citeauthoryear{EuroPOND~consortium}{EuroPOND~consortium}{[n.
  d.]}]%
        {tadpole}
\bibfield{author}{\bibinfo{person}{ADNI EuroPOND~consortium}.}
  \bibinfo{year}{[n. d.]}\natexlab{}.
\newblock \bibinfo{title}{TADPOLE Challenge}.
\newblock   (\bibinfo{year}{[n. d.]}).
\newblock
\urldef\tempurl%
\url{https://tadpole.grand-challenge.org/}
\showURL{%
\tempurl}


\bibitem[\protect\citeauthoryear{Figueredo and Wolf}{Figueredo and
  Wolf}{2009}]%
        {Figueredo:2009dg}
\bibfield{author}{\bibinfo{person}{A.~J. Figueredo} {and}
  \bibinfo{person}{P.~S.~A. Wolf}.} \bibinfo{year}{2009}\natexlab{}.
\newblock \showarticletitle{Assortative pairing and life history strategy - a
  cross-cultural study.}
\newblock \bibinfo{journal}{\emph{Human Nature}}  \bibinfo{volume}{20}
  (\bibinfo{year}{2009}), \bibinfo{pages}{317--330}.
\newblock


\bibitem[\protect\citeauthoryear{Greff, Srivastava, Koutn{\'\i}k, Steunebrink,
  and Schmidhuber}{Greff et~al\mbox{.}}{2017}]%
        {greff2017lstm}
\bibfield{author}{\bibinfo{person}{Klaus Greff}, \bibinfo{person}{Rupesh~K
  Srivastava}, \bibinfo{person}{Jan Koutn{\'\i}k}, \bibinfo{person}{Bas~R
  Steunebrink}, {and} \bibinfo{person}{J{\"u}rgen Schmidhuber}.}
  \bibinfo{year}{2017}\natexlab{}.
\newblock \showarticletitle{LSTM: A search space odyssey}.
\newblock \bibinfo{journal}{\emph{IEEE transactions on neural networks and
  learning systems}} (\bibinfo{year}{2017}).
\newblock


\bibitem[\protect\citeauthoryear{Hartigan and Wong}{Hartigan and Wong}{1979}]%
        {hartigan1979algorithm}
\bibfield{author}{\bibinfo{person}{John~A Hartigan} {and}
  \bibinfo{person}{Manchek~A Wong}.} \bibinfo{year}{1979}\natexlab{}.
\newblock \showarticletitle{Algorithm AS 136: A k-means clustering algorithm}.
\newblock \bibinfo{journal}{\emph{Journal of the Royal Statistical Society.
  Series C (Applied Statistics)}} \bibinfo{volume}{28}, \bibinfo{number}{1}
  (\bibinfo{year}{1979}), \bibinfo{pages}{100--108}.
\newblock


\bibitem[\protect\citeauthoryear{Hochreiter and Schmidhuber}{Hochreiter and
  Schmidhuber}{1997}]%
        {hochreiter1997long}
\bibfield{author}{\bibinfo{person}{Sepp Hochreiter} {and}
  \bibinfo{person}{J{\"u}rgen Schmidhuber}.} \bibinfo{year}{1997}\natexlab{}.
\newblock \showarticletitle{Long short-term memory}.
\newblock \bibinfo{journal}{\emph{Neural computation}} \bibinfo{volume}{9},
  \bibinfo{number}{8} (\bibinfo{year}{1997}), \bibinfo{pages}{1735--1780}.
\newblock


\bibitem[\protect\citeauthoryear{Khan}{Khan}{2014}]%
        {khan2014economic}
\bibfield{author}{\bibinfo{person}{Mohsin Khan}.}
  \bibinfo{year}{2014}\natexlab{}.
\newblock \bibinfo{booktitle}{\emph{The economic consequences of the Arab
  Spring}}.
\newblock \bibinfo{publisher}{Atlantic Council of the United States}.
\newblock


\bibitem[\protect\citeauthoryear{Kimoto, Asakawa, Yoda, and Takeoka}{Kimoto
  et~al\mbox{.}}{1990}]%
        {kimoto1990stock}
\bibfield{author}{\bibinfo{person}{Takashi Kimoto}, \bibinfo{person}{Kazuo
  Asakawa}, \bibinfo{person}{Morio Yoda}, {and} \bibinfo{person}{Masakazu
  Takeoka}.} \bibinfo{year}{1990}\natexlab{}.
\newblock \showarticletitle{Stock market prediction system with modular neural
  networks}. In \bibinfo{booktitle}{\emph{Neural Networks, 1990., 1990 IJCNN
  International Joint Conference on}}. IEEE, \bibinfo{pages}{1--6}.
\newblock


\bibitem[\protect\citeauthoryear{Kingma and Ba}{Kingma and Ba}{2014}]%
        {DBLP:journals/corr/KingmaB14}
\bibfield{author}{\bibinfo{person}{Diederik~P. Kingma} {and}
  \bibinfo{person}{Jimmy Ba}.} \bibinfo{year}{2014}\natexlab{}.
\newblock \showarticletitle{Adam: {A} Method for Stochastic Optimization}.
\newblock \bibinfo{journal}{\emph{CoRR}}  \bibinfo{volume}{abs/1412.6980}
  (\bibinfo{year}{2014}).
\newblock
\showeprint[arxiv]{1412.6980}
\urldef\tempurl%
\url{http://arxiv.org/abs/1412.6980}
\showURL{%
\tempurl}


\bibitem[\protect\citeauthoryear{Lee, Xie, Gallagher, Zhang, and Tu}{Lee
  et~al\mbox{.}}{2015}]%
        {lee2015deeply}
\bibfield{author}{\bibinfo{person}{Chen-Yu Lee}, \bibinfo{person}{Saining Xie},
  \bibinfo{person}{Patrick Gallagher}, \bibinfo{person}{Zhengyou Zhang}, {and}
  \bibinfo{person}{Zhuowen Tu}.} \bibinfo{year}{2015}\natexlab{}.
\newblock \showarticletitle{Deeply-supervised nets}. In
  \bibinfo{booktitle}{\emph{Artificial Intelligence and Statistics}}.
  \bibinfo{pages}{562--570}.
\newblock


\bibitem[\protect\citeauthoryear{Lewis and Trempe}{Lewis and Trempe}{2017}]%
        {tagkey2017iii}
\bibfield{author}{\bibinfo{person}{Thomas~J. Lewis} {and}
  \bibinfo{person}{Clement~L. Trempe}.} \bibinfo{year}{2017}\natexlab{}.
\newblock \showarticletitle{The Brain and Beyond}.
\newblock  (\bibinfo{year}{2017}), \bibinfo{pages}{4}.
\newblock
\showISBNx{978-0-12-812112-2}
\urldef\tempurl%
\url{https://doi.org/10.1016/B978-0-12-812112-2.00019-7}
\showDOI{\tempurl}


\bibitem[\protect\citeauthoryear{Lipton, Kale, Elkan, and Wetzell}{Lipton
  et~al\mbox{.}}{2015}]%
        {lipton2015learning}
\bibfield{author}{\bibinfo{person}{Zachary~C Lipton}, \bibinfo{person}{David~C
  Kale}, \bibinfo{person}{Charles Elkan}, {and} \bibinfo{person}{Randall
  Wetzell}.} \bibinfo{year}{2015}\natexlab{}.
\newblock \showarticletitle{Learning to diagnose with LSTM recurrent neural
  networks}.
\newblock \bibinfo{journal}{\emph{arXiv preprint arXiv:1511.03677}}
  (\bibinfo{year}{2015}).
\newblock


\bibitem[\protect\citeauthoryear{Mart{\'{\i}}n~Abadi
  et~al\mbox{.}}{Mart{\'{\i}}n~Abadi et~al\mbox{.}}{2016}]%
        {DBLP:journals/corr/AbadiABBCCCDDDG16}
\bibfield{author}{\bibinfo{person}{Ashish~Agarwal Mart{\'{\i}}n~Abadi}
  {et~al\mbox{.}}} \bibinfo{year}{2016}\natexlab{}.
\newblock \showarticletitle{TensorFlow: Large-Scale Machine Learning on
  Heterogeneous Distributed Systems}.
\newblock \bibinfo{journal}{\emph{CoRR}}  \bibinfo{volume}{abs/1603.04467}
  (\bibinfo{year}{2016}).
\newblock
\showeprint[arxiv]{1603.04467}
\urldef\tempurl%
\url{http://arxiv.org/abs/1603.04467}
\showURL{%
\tempurl}


\bibitem[\protect\citeauthoryear{Nestor, Rupsingh, Borrie, Smith, Accomazzi,
  Wells, Fogarty, Bartha, and Initiative}{Nestor et~al\mbox{.}}{2008}]%
        {nestor2008ventricular}
\bibfield{author}{\bibinfo{person}{Sean~M Nestor}, \bibinfo{person}{Raul
  Rupsingh}, \bibinfo{person}{Michael Borrie}, \bibinfo{person}{Matthew Smith},
  \bibinfo{person}{Vittorio Accomazzi}, \bibinfo{person}{Jennie~L Wells},
  \bibinfo{person}{Jennifer Fogarty}, \bibinfo{person}{Robert Bartha}, {and}
  \bibinfo{person}{Alzheimer's Disease~Neuroimaging Initiative}.}
  \bibinfo{year}{2008}\natexlab{}.
\newblock \showarticletitle{Ventricular enlargement as a possible measure of
  Alzheimer's disease progression validated using the Alzheimer's disease
  neuroimaging initiative database}.
\newblock \bibinfo{journal}{\emph{Brain}} \bibinfo{volume}{131},
  \bibinfo{number}{9} (\bibinfo{year}{2008}), \bibinfo{pages}{2443--2454}.
\newblock


\bibitem[\protect\citeauthoryear{NOAA}{NOAA}{2018}]%
        {NOAA}
\bibfield{author}{\bibinfo{person}{NOAA}.} \bibinfo{year}{2018}\natexlab{}.
\newblock \showarticletitle{NOAA National Centers for Environmental
  information, Climate at a Glance}.
\newblock \bibinfo{journal}{\emph{U.S. Time Series, Average Temperature
  http://www.ncdc.noaa.gov/cag/}} (\bibinfo{year}{2018}).
\newblock


\bibitem[\protect\citeauthoryear{Ott, Cohen, Okonkwo, Johanson, Stopa, Donahue,
  and Silverberg}{Ott et~al\mbox{.}}{2009}]%
        {ott2009relationship}
\bibfield{author}{\bibinfo{person}{Brian~R Ott}, \bibinfo{person}{Ronald~A
  Cohen}, \bibinfo{person}{Ozioma~C Okonkwo}, \bibinfo{person}{Conrad~E
  Johanson}, \bibinfo{person}{Edward~G Stopa}, \bibinfo{person}{John~E
  Donahue}, {and} \bibinfo{person}{Gerald~D Silverberg}.}
  \bibinfo{year}{2009}\natexlab{}.
\newblock \showarticletitle{The relationship between brain ventricular volume
  and cerebrospinal fluid levels of A-beta and tau in apolipoprotein E4
  positive normal controls and patients with Alzheimer's disease}.
\newblock \bibinfo{journal}{\emph{Alzheimer's \& Dementia: The Journal of the
  Alzheimer's Association}} \bibinfo{volume}{5}, \bibinfo{number}{4}
  (\bibinfo{year}{2009}), \bibinfo{pages}{P299--P300}.
\newblock


\bibitem[\protect\citeauthoryear{Outlook}{Outlook}{2008}]%
        {outlook2008energy}
\bibfield{author}{\bibinfo{person}{Annual~Energy Outlook}.}
  \bibinfo{year}{2008}\natexlab{}.
\newblock \showarticletitle{Energy Information Administration}.
\newblock \bibinfo{journal}{\emph{Official Energy Statistics from the US
  Government: www. eia. doe. gov}} (\bibinfo{year}{2008}).
\newblock


\bibitem[\protect\citeauthoryear{Rousseeuw}{Rousseeuw}{1987}]%
        {rousseeuw1987silhouettes}
\bibfield{author}{\bibinfo{person}{Peter~J Rousseeuw}.}
  \bibinfo{year}{1987}\natexlab{}.
\newblock \showarticletitle{Silhouettes: a graphical aid to the interpretation
  and validation of cluster analysis}.
\newblock \bibinfo{journal}{\emph{Journal of computational and applied
  mathematics}}  \bibinfo{volume}{20} (\bibinfo{year}{1987}),
  \bibinfo{pages}{53--65}.
\newblock


\bibitem[\protect\citeauthoryear{Wu, Ahmed, Beutel, Smola, and Jing}{Wu
  et~al\mbox{.}}{2017}]%
        {RRN}
\bibfield{author}{\bibinfo{person}{Chao-Yuan Wu}, \bibinfo{person}{Amr Ahmed},
  \bibinfo{person}{Alex Beutel}, \bibinfo{person}{Alexander~J Smola}, {and}
  \bibinfo{person}{How Jing}.} \bibinfo{year}{2017}\natexlab{}.
\newblock \showarticletitle{Recurrent recommender networks}. In
  \bibinfo{booktitle}{\emph{Proceedings of the Tenth ACM International
  Conference on Web Search and Data Mining}}. ACM, \bibinfo{pages}{495--503}.
\newblock


\end{thebibliography}

\end{document}